\documentclass[conference]{IEEEtran}
\IEEEoverridecommandlockouts
\usepackage{cite}
\usepackage{float}
\usepackage{amsmath,amssymb,amsfonts}
\usepackage{algorithmic}
\usepackage{graphicx}
\usepackage{textcomp}
\usepackage[dvipsnames]{xcolor}
\usepackage{tikz}
\usetikzlibrary{positioning}
\usepackage{subcaption}

\def\BibTeX{{\rm B\kern-.05em{\sc i\kern-.025em b}\kern-.08em
    T\kern-.1667em\lower.7ex\hbox{E}\kern-.125emX}}
\begin{document}

\title{Fuzzy Theory in Computer Vision: A Review}

% Fuzzy-Based Approaches in Computer Vision: A Review of Theory and Applications
\author{\IEEEauthorblockN{Adilet Yerkin, Ayan Igali, Elnara Kadyrgali, Maksat Shagyrov, Malika Ziyada, \\Muragul Muratbekova, Pakizar Shamoi\IEEEauthorrefmark{1}}
\IEEEauthorblockA{School of Information Technology and Engineering \\
Kazakh-British Technical University\\
Almaty, Kazakhstan\\
Email: 
\IEEEauthorrefmark{1}p.shamoi@kbtu.kz
}
}

\maketitle

\begin{abstract}

Computer vision applications are omnipresent nowadays. The current paper explores the use of fuzzy logic in computer vision, stressing its role in handling uncertainty, noise, and imprecision in image data. Fuzzy logic is able to model gradual transitions and human-like reasoning and provides a promising approach to computer vision. Fuzzy approaches offer a way to improve object recognition, image segmentation, and feature extraction by providing more adaptable and interpretable solutions compared to traditional methods. We discuss key fuzzy techniques, including fuzzy clustering, fuzzy inference systems, type-2 fuzzy sets, and fuzzy rule-based decision-making.  The paper also discusses various applications, including medical imaging, autonomous systems, and industrial inspection. Additionally, we explore the integration of fuzzy logic with deep learning models such as convolutional neural networks (CNNs) to enhance performance in complex vision tasks. Finally, we examine emerging trends such as hybrid fuzzy-deep learning models and explainable AI.

\end{abstract}


\section{Introduction}

% 1-2 paragraphs fairy tale - very general. Computer vision simple definitions, need for imprecision.

Computer vision is a subfield of artificial intelligence focused on analyzing and interpreting visual data from images and videos \cite{CV2002}. It includes multiple levels of processing, including low-level tasks such as noise removal and edge detection, intermediate-level tasks like segmentation and object detection, and high-level tasks such as scene understanding and behavior recognition \cite{review2003}. 

The ultimate goal of computer vision is to replicate human perception by processing and interpreting visual input in a way that aligns with how humans perceive the world. However, traditional methods often struggle with uncertainty, noise, and imprecision in visual data \cite{1676732}.

Fuzzy logic improves computer vision systems by introducing human-like reasoning to handle uncertainty, imprecision, noise, and gradual transitions in visual data \cite{review2003}. Unlike traditional methods that rely on crisp boundaries, fuzzy logic allows for soft classifications and adaptive processing \cite{SHARMA2023100751}, making it particularly useful in complex visual environments. As a result, vision models interpret images with degrees of belonging rather than rigid binary labels, closely mimicking human perception.

Fuzzy approaches make object recognition and detection more adaptable and interpretable \cite{Połap2021Image}, which is useful in areas like medical imaging, autonomous driving, and industrial inspection. This allows the creation of more reliable and human-like decision-making systems in computer vision.

% Computer vision is a unit of artificial intelligence that covers the analysis of various features within a given input image or video \cite{CV2002}. It involves low-level (noise removal, edge detection, etc.), intermediate-level (segmentation, object detection, etc.), and high-level (scene understanding, behaviour recognition, etc.) vision algorithms to reach the full vision of input data. The primary objective of CV is to replicate human perception and process input data in the same way we do; therefore, how can fuzzy logic assist in achieving this goal?

% 1-2 paragraphs - why fuzzy logic is a good choice  in addressing these challenges and it is popular among researchers.

% \colorbox{LimeGreen}{7.
% Research justification - why important to study (Pakita) }

% \colorbox{Goldenrod}{5.
% Main message}

This paper aims to provide a review and bibliometric analysis of fuzzy-based approaches in computer vision, emphasizing their role in handling uncertainty, noise, and imprecision in image analysis. Furthermore, the paper discusses the significance of fuzziness in aligning computer vision algorithms with human perception, increasing interpretability and adaptability in real-world applications.

While there are a few existing review articles on fuzzy logic in computer vision, they are either outdated (e.g., from 1992 \cite{krishnapuram1992fuzzy} and 2003 \cite{review2003}) or focusing on specific applications such as medical image processing \cite{baimukashev2024systematic}. A notable review \cite{review2003} from 2003 provides valuable insights. However, our work offers a broader range of applications and approaches, reflecting the significant advancements in the field over the past two decades. By conducting a comprehensive bibliometric analysis of over 4,000 publications, we provide a more up-to-date and extensive overview of fuzzy logic's role in computer vision, capturing new trends and emerging research areas.

% \textcolor{Fuchsia}{... discuss fuzzy-based approaches in computer vision techniques and the significance of fuzziness in aligning computer vision algorithms with the human visual process.}

The paper has the following contributions:
\begin{itemize}
\item  Review of fuzzy-based computer vision techniques, including fuzzy clustering, fuzzy inference systems, type-2 fuzzy sets, and hybrid models
\item Bibliometric analysis of research trends
    \item Analysis of Real-World Applications
    
    \item  Identification of emerging trends, including explainable AI, type-2 fuzzy sets, fuzzy quantum computing, and neuro-fuzzy learning. Discussion of open challenges, such as computational efficiency and hybrid fuzzy-AI integration.
    
    \item Analysis over 4000 publications to highlight current trends and advancements in fuzzy logic within computer vision.

    \item An in-depth overview of seven key fuzzy-based techniques and their implementation across five distinct applications, enhancing understanding of their practical impact.

\end{itemize}

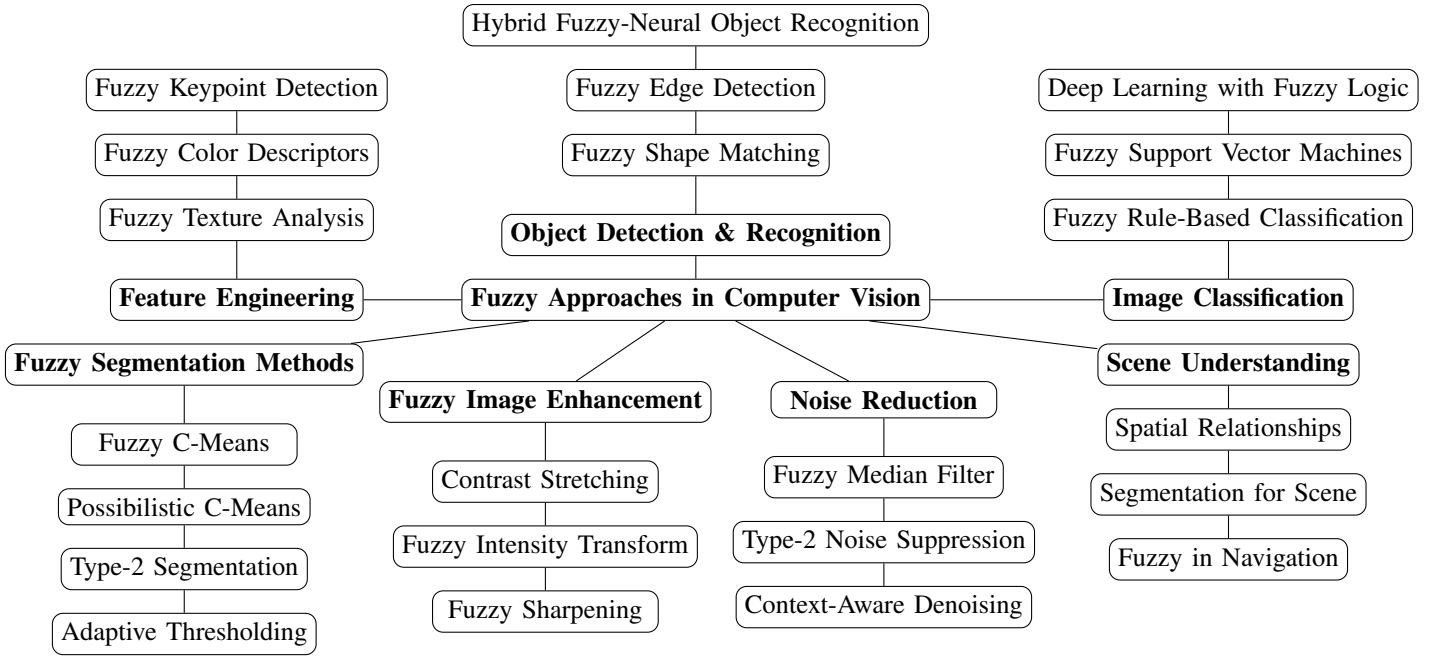
\begin{figure*}[ht!]
    \centering
    \begin{tikzpicture}
        [
            node distance=0.3cm and 1.3cm,
            every node/.style={draw, text centered, minimum width=3cm, rounded corners},
            main/.style={rectangle, draw, minimum width=4cm, font=\bfseries},
            branch/.style={rectangle, draw, minimum width=3cm}
        ]
        
        \node[main] (A) {Fuzzy Approaches in Computer Vision};
        
        \node[branch, below left=of A] (B) {\textbf{Fuzzy Segmentation Methods}};
        \node[branch, below=of A, xshift=-2cm, yshift=-0.5cm] (C) {\textbf{Fuzzy Image Enhancement}};
        \node[branch, below=of A, xshift=2.5cm, yshift=-0.5cm] (D) {\textbf{Noise Reduction}};
        
        \draw (A) -- (B);
        \draw (A) -- (C);
        \draw (A) -- (D);
        
        \node[branch, above=of A] (E) {\textbf{Object Detection \& Recognition}};
        \node[branch, right=of A, xshift=1cm] (F) {\textbf{Image Classification}};
        \node[branch, left=of A] (G) {\textbf{Feature Engineering}}; 
        
        \draw (A) -- (E);
        \draw (A) -- (G);
        \draw (A) -- (F);
        
        \node[branch, above=of G, yshift=0.2cm] (G1) {Fuzzy Texture Analysis};
        \node[branch, above=of G1] (G2) {Fuzzy Color Descriptors};
        \node[branch, above=of G2] (G3) {Fuzzy Keypoint Detection};
        
        \draw (G) -- (G1);
        \draw (G1) -- (G2);
        \draw (G2) -- (G3);
        
        \node[branch, above=of E, yshift=0.2cm] (E1) {Fuzzy Shape Matching};
        \node[branch, above=of E1] (E2) {Fuzzy Edge Detection};
        \node[branch, above=of E2] (E3) {Hybrid Fuzzy-Neural Object Recognition};
        
        \draw (E) -- (E1);
        \draw (E1) -- (E2);
        \draw (E2) -- (E3);
        
        \node[branch, above=of F, yshift=0.2cm] (F1) {Fuzzy Rule-Based Classification};
        \node[branch, above=of F1] (F2) {Fuzzy Support Vector Machines};
        \node[branch, above=of F2] (F3) {Deep Learning with Fuzzy Logic};
        
        \draw (F) -- (F1);
        \draw (F1) -- (F2);
        \draw (F2) -- (F3);
        
        \node[branch, below=of F] (I) {\textbf{Scene Understanding}};
        
        \draw (A) -- (I);

        \node[branch, below= of B, yshift=-0.2cm] (B1) {Fuzzy C-Means};
        \node[branch, below= of B1] (B2) {Possibilistic C-Means};
        \node[branch, below= of B2] (B3) {Type-2 Segmentation};
        \node[branch, below= of B3] (B4) {Adaptive Thresholding};

        \draw (B) -- (B1);
        \draw (B1) -- (B2);
        \draw (B2) -- (B3);
        \draw (B3) -- (B4);

        \node[branch, below=of C, yshift=-0.2cm] (C1) {Contrast Stretching};
        \node[branch, below=of C1] (C2) {Fuzzy Intensity Transform};
        \node[branch, below=of C2] (C3) {Fuzzy Sharpening};

        \draw (C) -- (C1);
        \draw (C1) -- (C2);
        \draw (C2) -- (C3);

        \node[branch, below=of D, yshift=-0.2cm] (D1) {Fuzzy Median Filter};
        \node[branch, below=of D1] (D2) {Type-2 Noise Suppression};
        \node[branch, below=of D2] (D3) {Context-Aware Denoising};

        \draw (D) -- (D1);
        \draw (D1) -- (D2);
        \draw (D2) -- (D3);

        \node[branch, below= of I] (I1) {Spatial Relationships};
        \node[branch, below= of I1] (I2) {Segmentation for Scene};
        \node[branch, below= of I2] (I3) {Fuzzy in Navigation};

        \draw (I) -- (I1);
        \draw (I1) -- (I2);
        \draw (I2) -- (I3);
        
    \end{tikzpicture}
    \caption{Fuzzy Approaches in Computer Vision}
    \label{fig:fuzzy_approaches}
\end{figure*}

 % \textcolor{purple}{we added 1 more section}
The paper has the following structure. Section I introduces the topic. Section II provides an overview of the fundamentals of fuzzy logic, including fuzzy set theory, membership functions, and fuzzy inference systems. Section III discusses various fuzzy approaches in computer vision, including segmentation, object detection, noise reduction, feature extraction, and classification, as well as their integration with deep learning. Section IV explores applications of fuzzy-based computer vision. Next, Section V presents the biometric analysis. Finally, Section VI provides concluding remarks.

\section{Fundamentals of Fuzzy Logic}
% Fuzzy logic vs. traditional and deep learning-based approaches. Tipa we have DL as well. We need to mention it and explain adv. of fuzzy. Motivation for using fuzzy in CV and fuzzy basics. \textcolor{Fuchsia}{Eto I v section Trends est, da?}

While deep learning has achieved remarkable success in computer vision \cite{Voulodimos2018}, it often requires large datasets and extensive computational resources and lacks interpretability. In contrast, fuzzy logic offers a human-like reasoning framework that effectively handles uncertainty, imprecision, and limited data. Computer vision systems can achieve greater robustness, adaptability, and explainability by integrating fuzzy approaches with traditional and deep learning-based methods.

Fuzzy set theory, introduced by Zadeh in 1965 \cite{zadeh1965fuzzy}, extends classical set theory by allowing elements to have partial membership rather than a strict binary classification (0 or 1).

\newtheorem{definition}{Definition}

\begin{definition}
Let \( \mathcal{X} \) be a universe of discourse. A \textit{fuzzy set} \( A \) in \( \mathcal{X} \) is defined by a \textit{membership function} \cite{zadeh1965fuzzy}:
\begin{equation}
    \mu_A: \mathcal{X} \to [0,1],
\end{equation}
which assigns to each element \( x \in \mathcal{X} \) a membership degree \( \mu_A(x) \), indicating the degree to which \( x \) belongs to \( A \). The fuzzy set \( A \) is formally represented as:
\begin{equation}
    A = \{(x, \mu_A(x)) \mid x \in \mathcal{X} \}.
\end{equation}
\end{definition}

\begin{definition}
A \textit{linguistic variable} \( \mathcal{V} \) is defined as a quintuple \cite{zadeh1975concept}:
\begin{equation}
    \mathcal{V} = (N, T, \mathcal{U}, S, M),
\end{equation}
where:  
- \( N \) is the name of the variable,  
- \( T \) is a finite set of linguistic terms the variable can take,  
- \( \mathcal{U} \) is the universe of discourse over which the linguistic terms are defined,  
- \( S \) is a syntactic rule for generating linguistic terms,  
- \( M \) is a semantic rule that maps each linguistic term \( L_i \in T \) to a fuzzy set in \( \mathcal{U} \).
\end{definition}

Each linguistic term is associated with a membership function \( \mu_{L_i}(u) \), which defines the degree to which a given \( u \in \mathcal{U} \) satisfies the linguistic term.

Fuzzy rules are a key component of fuzzy logic systems \cite{Zadeh1988}. Fuzzy rules are used to encode expert knowledge and allow the system to make inferences about data in a way that human reasoning. A fuzzy rule is a simple IF-THEN statement where both the IF part (the antecedent) and the THEN part (the consequent) contain fuzzy sets. For instance, the following fuzzy rules illustrate how attributes such as color, shape, brightness, and contrast can be utilized for object classification:
\begin{align*}
\text{Rule 1: } & \text{IF } \text{Color is Green} \text{ AND } \text{Shape is Irregular} \\
                & \text{THEN } \text{ObjectClass is Likely Vegetation}. \\
\text{Rule 2: } & \text{IF } \text{Brightness is Low} \text{ AND } \text{Contrast is High} \\
                & \text{THEN } \text{ObjectClass is Likely a Shadow}.
\end{align*}

The concept of a fuzzy inference system (FIS) was developed within the broader field of fuzzy logic. A fuzzy inference system is a framework for processing data and making decisions based on fuzzy rules that can handle imprecision and uncertainty.

\section{Methodology}

We began our review by conducting a thorough search for relevant literature across two widely used academic databases: Scopus and Google Scholar. Our search focused on a set of core keywords that reflect the main themes of this study, including "fuzzy sets, "fuzzy logic," and "computer vision." To keep the scope broad and inclusive, we did not apply any filters related to publication date or document type and collected 66 papers for review that matched all search criteria. The screening process was carried out in two phases: we first examined the titles and abstracts to identify potentially relevant studies, followed by a manual review of full texts to confirm the papers' relevance and depth of contribution to the intersection of fuzzy theory and computer vision. This structured approach allowed us to focus on the most insightful and meaningful studies, providing a solid base for our analysis and discussion. Fig. \ref{fig:fuzzy_approaches} provides visual support. Then we combined findings to keep a classified vision of all fuzzy approaches and applications in Sections IV and  V.

The main analysis part includes a collection of the dataset using the search query "Fuzzy computer vision", which includes 4,305 documents from Scopus published between 1973 and 2025. We used the full timeframe of existing papers available in the database to keep a complete record of research findings. Table \ref{tab:top_publishers} presents information about the number of papers on fuzzy computer vision by publishers. IEEE emerges as the leading publisher in the topic (961), followed by Springer (494) and Elsevier (248).

Afterwards, we conducted a bibliometric analysis of publications on fuzzy computer vision to express publication trends over time and key findings.

\begin{table}[h]
\centering
\caption{Top 10 Publishers by Number of Publications}
\label{tab:top_publishers}
\begin{tabular}{|l|c|}
\hline
                                        \textbf{Publisher} &  \textbf{Publication Count} \\
\hline
                              IEEE &                961 \\
                          Springer &                494 \\
                          Elsevier &                248 \\
                              SPIE &                 137 \\
                              MDPI &                 78 \\
                             Wiley &                 34 \\
                   Hindawi Limited &                 29 \\
Association for Computing Machinery &                 29 \\
IOS Press BV &                 21 \\
Institute of Physics Publishing & 20 \\
\hline
\end{tabular}
\end{table}

\section{Fuzzy Approaches in Computer Vision}

% Fuzzy approaches in computer vision provide a robust framework for handling uncertainty and imprecision in image analysis. By leveraging fuzzy logic, these methods enhance segmentation, object recognition, and scene understanding, enabling more adaptive and human-like perception in complex visual environments.

In this section, we provide a detailed overview of various fuzzy-based techniques and their applications in different levels of visual processing. 

% \textcolor{purple}{ we moved it to methods --- Figure \ref{fig:fuzzy_approaches} provides visual support.}

% \colorbox{LimeGreen}{15. Figure with approaches (Muragul) -- REVIEW PLS}

\subsection{Fuzzy Segmentation Methods}
\subsubsection{Fuzzy C-Means (FCM) and its extensions}

Fuzzy C-Means (FCM) clustering is widely used in computer vision for handling uncertainty in image segmentation and pattern recognition. Unlike hard clustering methods, FCM assigns membership probabilities to data points, allowing for more flexible segmentation, particularly in complex visual environments \cite{Liu2019, Lei2019, Wang2021}. This makes FCM effective for tasks like object recognition, medical imaging, and scene analysis, where boundaries may be unclear.
Recent advancements focus on improving segmentation accuracy and efficiency. Superpixel-based FCM preserves local spatial structures, reducing computational complexity while enhancing segmentation performance \cite{Lei2019}. Residual-driven FCM incorporates noise estimation and spatial constraints, improving robustness in medical and real-world images \cite{Wang2021}.

Hybrid models combining FCM with superpixel segmentation, Random Forest classifiers, and artificial neural networks (ANNs) further enhance object and behavior recognition, benefiting applications in surveillance, autonomous driving, and smart healthcare \cite{Jalal2021}.

\subsubsection{Type-2 fuzzy sets for image segmentation}
Interval Type-2 Fuzzy Sets (IT2FS) have shown great potential in image segmentation \cite{Kim2015, Zhao2019, Mishro2021, Ali2022}. Unlike T1FS, which uses crisp membership values, IT2FS incorporates an additional degree of freedom by defining membership functions with upper and lower bounds. This flexibility is particularly beneficial in image segmentation, where variations in lighting, noise, and object boundaries often lead to ambiguous pixel classifications. 

% By leveraging IT2FS, segmentation algorithms can achieve more robust and adaptive performance, leading to smoother region boundaries and improved differentiation of objects in complex visual scenes.

\subsection{Fuzzy-Based Object Detection and Recognition }
% - Pakita: Rule-based fuzzy systems for object recognition, describe how rules are usually obtained: they are learned and generated from data, subjectively defined, experts derived, etc...

Object detection using fuzzy techniques is crucial as it addresses the limitations of traditional methods, which often struggle with ambiguity and variability in visual data. These systems rely on a set of if-then fuzzy rules to map input image features to object categories. The performance of such systems depends on how the rules are obtained \cite{ai4010013}. Typically, rules can be derived from expert knowledge, data, or defined subjectively.

% Fuzzy logic provides a robust framework for handling uncertainty, enhancing the accuracy and adaptability of detection systems in complex environments. 

Recent advancements in computer vision using fuzzy logic have significantly improved object detection and recognition in complex environments. \cite{Tan_5569091} and \cite{Panda_7321774} leverage fuzzy logic for robust vehicle and moving object detection against dynamic backgrounds, demonstrating improved accuracy and reduced false positives. Studies such as \cite{Reyes_1189896} and \cite{Susan_6658004} apply these techniques to robot recognition \cite{Reyes_1189896}  and object classification \cite{Susan_6658004} using fuzzy-graph theoretical clustering and Gabor Wavelets, respectively, to effectively handle environmental variations. Meanwhile, works  \cite{Chacon_8070318}, \cite{Das_8256679}, \cite{Chacon_5686927},  \cite{Mahapatra_6622397}, and \cite{Zhou_8375812}  demonstrate the application of fuzzy logic in dynamic surveillance systems \cite{Chacon_8070318}, comprehensive background subtraction \cite{Das_8256679}, adaptive surveillance without manual tuning  \cite{Chacon_5686927}, human detection \cite{Mahapatra_6622397}, and saliency mapping  \cite{Zhou_8375812}, illustrating the broad utility and adaptability of fuzzy systems in real-time scenarios across various conditions and settings. Next, \cite{Juang5892887} proposed a novel real-time object detection system using a Takagi-Sugeno fuzzy system combined with a support vector machine in principal component space, which enhances detection accuracy through innovative global-local color feature analysis and efficient histogram computation.

% -Integration of fuzzy logic with CNNs for object detection.

\subsection{Fuzzy Image Enhancement}

Image enhancement is a key research area focused on improving visual quality and extracting information. Thus, several studies have explored the effectiveness of fuzzy-based approaches in image fusion problems \cite{yue2013fuzzy}, \cite{zhu2017fusion}, \cite{yang2016multimodal}. For instance, the authors in \cite{jiang2017novel} applied fuzzy logic to extract high-quality pixels, enabling the fusion of source images into a single, enhanced image. Applying stationary wavelet transform (SWT), the source images were decomposed into sub-images and transformed into fuzzy sets using the Gaussian membership function (GMF). Subsequently, the authors used local spatial frequency (LSF) to extract local features and developed a rule based on the fuzzy sets.

% Finally, they achieved a high-quality fused image by reconstructing it using the SWT. 

\subsection{Noise Reduction using Fuzzy Filtering}
Digital imaging specialists often face challenges with corrupted images \cite{sahu2012survey}. While earlier studies relied on simple median filters to address this issue, modern approaches now apply fuzzy logic controllers \cite{gupta2018study}. Authors in \cite{tavassoli2010new} proposed a method for impulse noise reduction using the fuzzy approach. They used the adaptive neuro-fuzzy inference system (ANFIS) to detect noisy pixels and fuzzy wavelet shrinkage (FWS) to change only the noise pixels for further filtering.

\subsection{Fuzzy Approach in Image Classification}

% -Just fuzzy?

% -Hybrid fuzzy-neural network models , ANFIS

Recent developments in image classification emphasize the incorporation of fuzzy logic to significantly enhance both the precision and adaptability of these systems across various applications.

Studies  \cite{shackelford2003hierarchical}, \cite{Kersten_1396324}, and \cite{Fauvel_1704969} demonstrate various uses of fuzzy logic in image classification: improving the classification of land cover in satellite images \cite{shackelford2003hierarchical}, improving land use classification through fuzzy clustering in SAR image analysis \cite{Kersten_1396324}, and refining urban classification with high-resolution imagery addressing pixel ambiguity \cite{Fauvel_1704969}. In \cite{Deng_7482843}, it is explored how fuzzy logic can enhance deep learning models to handle data uncertainty effectively, while in \cite{Chia_1232222}, it applies fuzzy logic to optimize land-use classification from SAR data. Next, \cite{Wang_46698} highlights the application of fuzzy supervised classification to geographical data, improving accuracy with sophisticated pixel representation.

Studies \cite{Cao_5893934} and \cite{Bardossy_992798} show how fuzzy logic facilitates complex image classifications in medical diagnostics \cite{Cao_5893934} and LANDSAT imagery \cite{Bardossy_992798}, respectively.

Next, authors in \cite{Moustakidis_5976442}, \cite{Margarit_5735208}, and \cite{Hung_5658101} illustrate the application of fuzzy logic in various classification frameworks: using fuzzy decision trees for forest images \cite{Moustakidis_5976442}, fuzzy neural networks for urban images \cite{Margarit_5735208}, and weighted fuzzy C-Means for high-dimensional pattern recognition \cite{Hung_5658101}. In \cite{Chanussot_1576686} demonstrates the use of derivative morphological profiles for urban area classification, utilizing fuzzy measurements to analyze local geometric information.

 \subsection{Fuzzy Feature Engineering}
Feature selection is critical for machine learning algorithms in computer vision \cite{8375658}. Fuzzy logic helps handle uncertainty in data, and feature extraction algorithms further improve classification performance \cite{DAS2020100288}.

Fuzzy edge detection enhances traditional methods by using a fuzzy rule-based evaluation to classify edge pixels with varying degrees of membership, improving edge detection in noisy, low-contrast images \cite{Raheja2021}. Fuzzy texture analysis refines feature extraction by applying fuzzy transformations to statistical descriptors, where a fuzzy-transformed co-occurrence vector enhances object detection and segmentation in dynamic backgrounds \cite{6236012}. Fuzzy color segmentation classifies pixels into color regions using fuzzy logic, improving adaptive segmentation over threshold-based methods. A fuzzy system for skin detection assigns degrees of skin-likeness to pixels, enhancing face detection accuracy under varying lighting conditions \cite{e19010026}.

Fuzzy feature selection helps reduce irrelevant and redundant features in computer vision tasks by assigning degrees of relevance rather than making binary inclusion/exclusion decisions. Unlike traditional feature selection methods, which may discard useful features due to rigid thresholds, fuzzy approaches provide a more adaptive selection process by considering feature importance with soft boundaries. A fuzzy feature selection method effectively selects useful features while controlling redundancy, leading to improved classification accuracy in both synthetic and real-world datasets \cite{7888460}. 

% This adaptability makes fuzzy feature selection particularly valuable in high-dimensional vision tasks, where conventional methods often struggle with feature dependencies and noise.

% write about: Fuzzy logic, sets in feature selection and dimensionality reduction.

\subsection{Scene Understanding}
% human activity recognition, behavior
% \textcolor{Tan}{
Scene understanding is a fundamental challenge in computer vision, requiring systems to segment, classify, and interpret objects and actions within a given environment. Human activity recognition (HAR) extends this concept by analyzing motion patterns and behaviors to understand interactions and intent \cite{Banerjee2015}. Fuzzy clustering methods, probabilistic models, and deep learning techniques are often combined to address the inherent uncertainty in dynamic scenes, where lighting conditions, occlusions, and object relationships vary over time. A combination of Maximum Entropy (ME), FCM, and ANN-based classification is employed to improve scene parsing and behavior recognition accuracy \cite{Jalal2021}. These approaches are particularly valuable in autonomous systems, intelligent surveillance, and assistive technologies, where precise scene interpretation and behavior analysis are essential for real-time decision-making.

\section{Applications}

% \textcolor{red}{We need to think how to represent it}
% \textcolor{red}{Integration with Deep learning}
% Advantages and Limitations of Fuzzy Logic in CV

% \textcolor{Fuchsia}{
% }
% Trends include mostly hybrid systems:
% -Fuzzy deep learning models

% -Type-2 and Intuitionistic Fuzzy Sets

% -Fuzzy Logic in Explainable AI (XAI) for interpretable computer vision models

% -Quantum Fuzzy Systems 

% -Fuzzy Machine Learning

\begin{figure}[tb]
    \centering
    \includegraphics[width=0.5\textwidth]{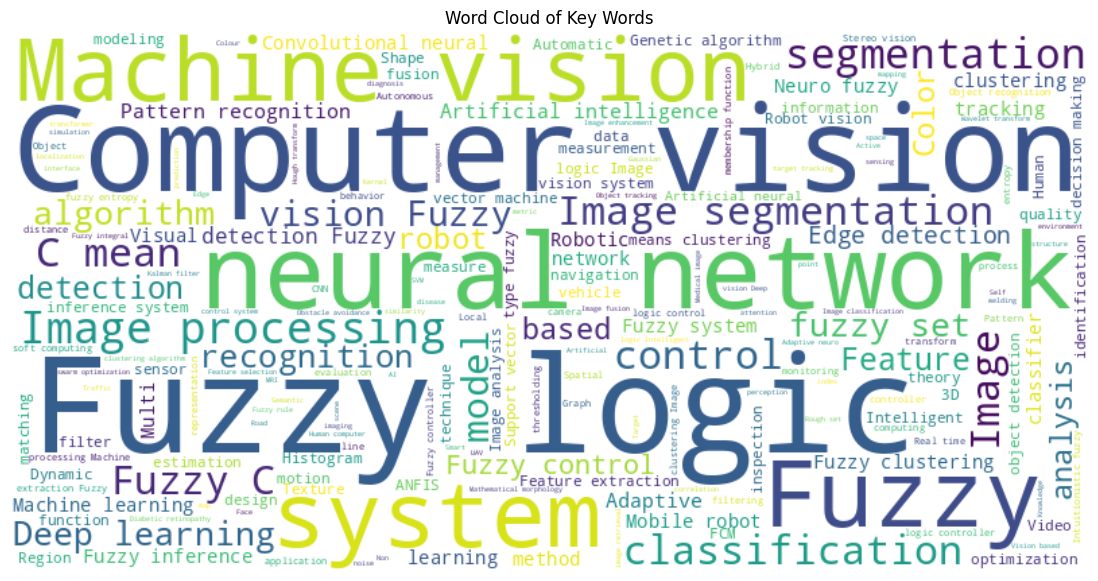}
    \caption{Word Cloud of Keywords}
    \label{fig_cloud}
\end{figure}

\begin{figure}[tb]
    \centering
    \includegraphics[width=0.5\textwidth]{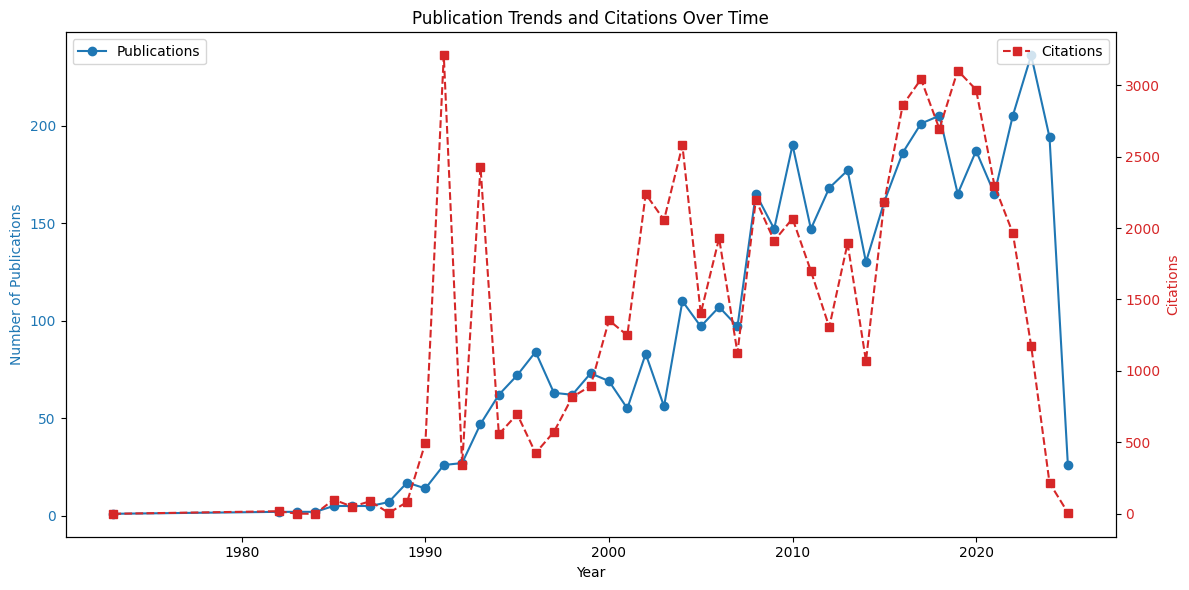}
    \caption{Publication Trends and Citations Over Time}
    \label{fig_trend_cite}
\end{figure}

\begin{figure}[tb]
    \centering
    % first subfigure
    \begin{subfigure}[b]{0.5\textwidth}
        \centering
        \includegraphics[width=\textwidth]{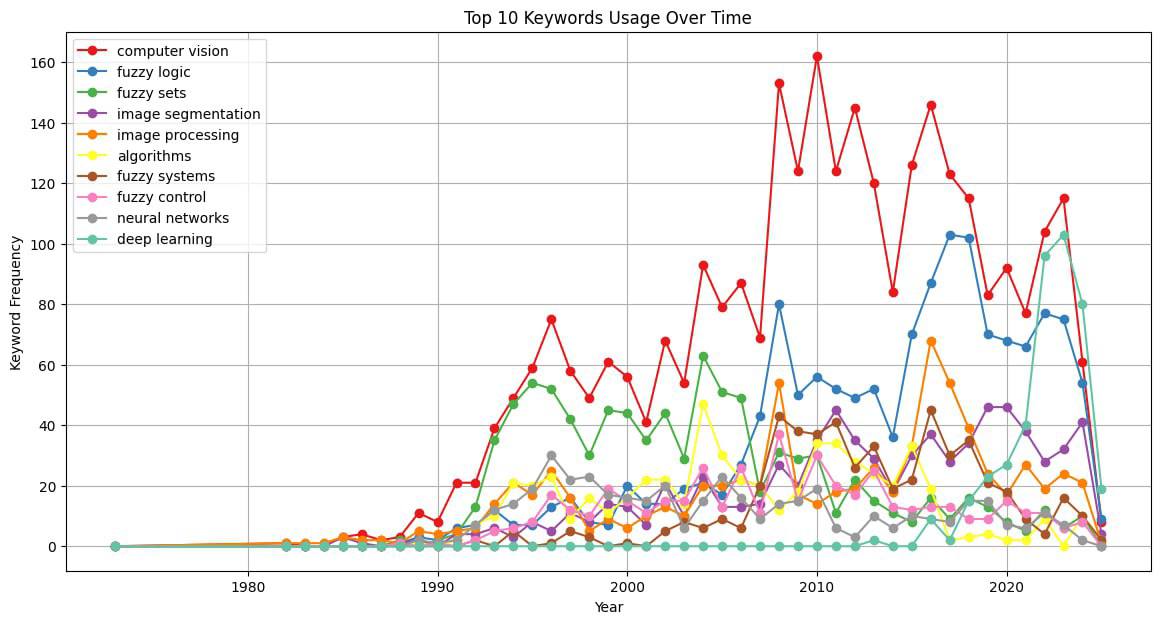}
        \caption{}
        \label{fig:overall_top10_key_sub1}
    \end{subfigure}

    \vspace{1em} % vertical spacing between subfigures

    % second subfigure
    \begin{subfigure}[b]{0.5\textwidth}
        \centering
        \includegraphics[width=\textwidth]{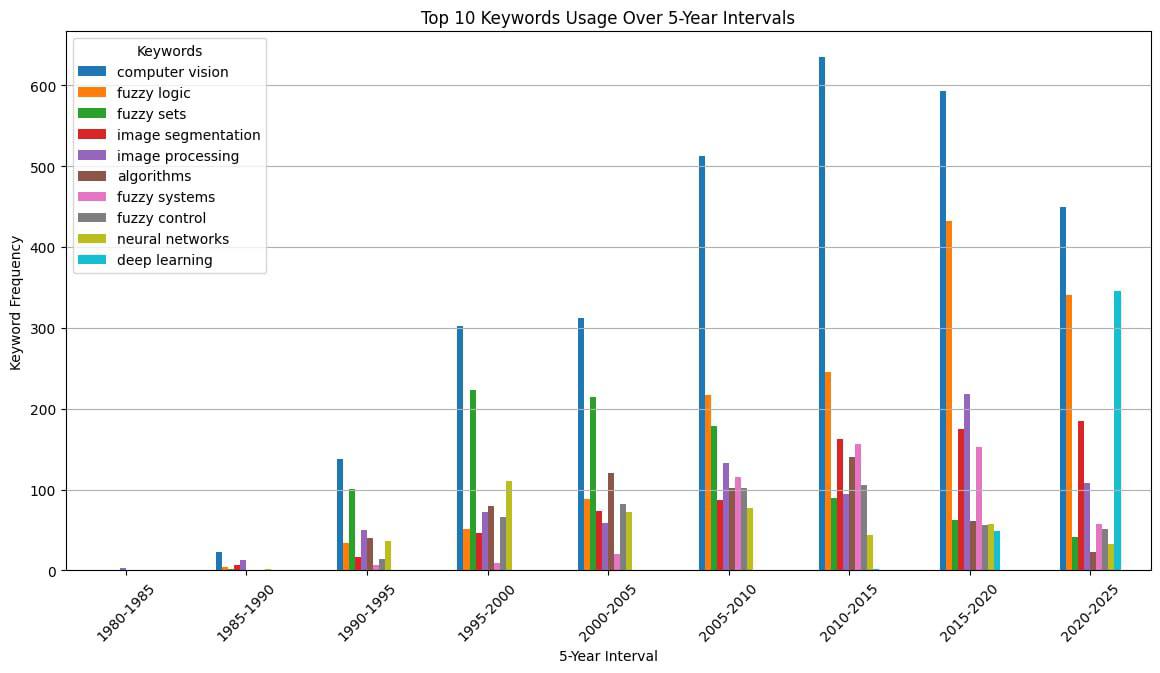}
        \caption{}
        \label{fig:overall_top10_key_sub2}
    \end{subfigure}

    \caption{Top 10 Keywords Trends in publications on fuzzy computer vision.}
    \label{fig:overall_top10_key}
\end{figure}

\begin{figure*}[tb]
    \centering
    \includegraphics[width=0.7\textwidth]{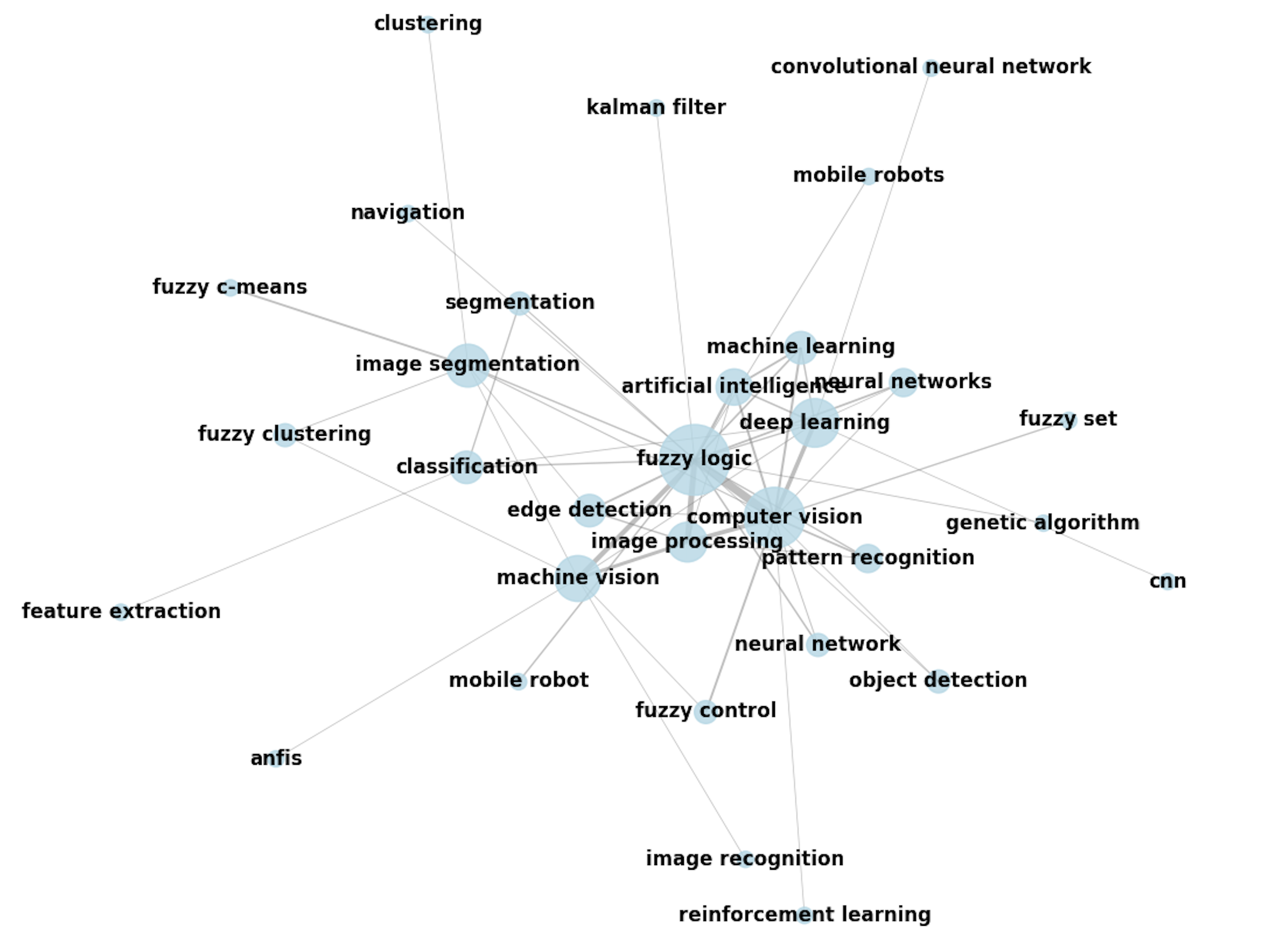}
    \caption{Keyword Co-occurrence Network}
    \label{fig_keyw_net}
\end{figure*}

Main applications include:
\begin{itemize}
    \item \textit{Robotic vision-control.} Fuzzy logic enhances robotic adaptability in uncertain environments. Type-2 fuzzy sets, in particular, may offer advantages over type-1 in handling dynamic conditions \cite{he2017adaptive, hagras2004hierarchical}. 
    \item \textit{Medical Image Analysis.} In medical diagnostics, combining CNNs with fuzzy logic can improve disease classification and treatment recommendations \cite{sharma2021deep}. Additionally, integrating fuzzy logic with vision transformers enables large-scale medical data analysis \cite{li2024fuzzy}.
    \item \textit{Fuzzy-based Affective Computing.} Given the fundamental role of emotions in human behavior \cite{pessoa2009emotion}, fuzzy logic can be utilized to enhance emotion recognition by integrating facial and speech data, thereby mitigating ambiguities in affective analysis \cite{wu2020two}.
    \item \textit{Remote Sensing, Satellite Images.} Fuzzy logic improves satellite image classification, particularly in complex urban environments \cite{shackelford2003hierarchical}. Neuro-fuzzy models may also support risk assessment and land-use management \cite{pradhan2010landslide}.
    \item \textit{Surveillance.} In surveillance systems, fuzzy logic can address uncertainties in object tracking, lighting variations, and scene interpretation, improving reliability \cite{10.1145/3444693}.
\end{itemize}

\section{Analysis and Results}
\label{sec:results}
% May be some sections (like trends) will be moved here. I am lost a bit

% Comparison of fuzzy-based methods vs. traditional/deep learning techniques - bibliometric analysis,accuracies or other metrics

% word clouds?

\begin{figure*}[tb]
    \centering
    \begin{subfigure}[t]{0.48\linewidth}
        \centering
        \includegraphics[width=\linewidth]{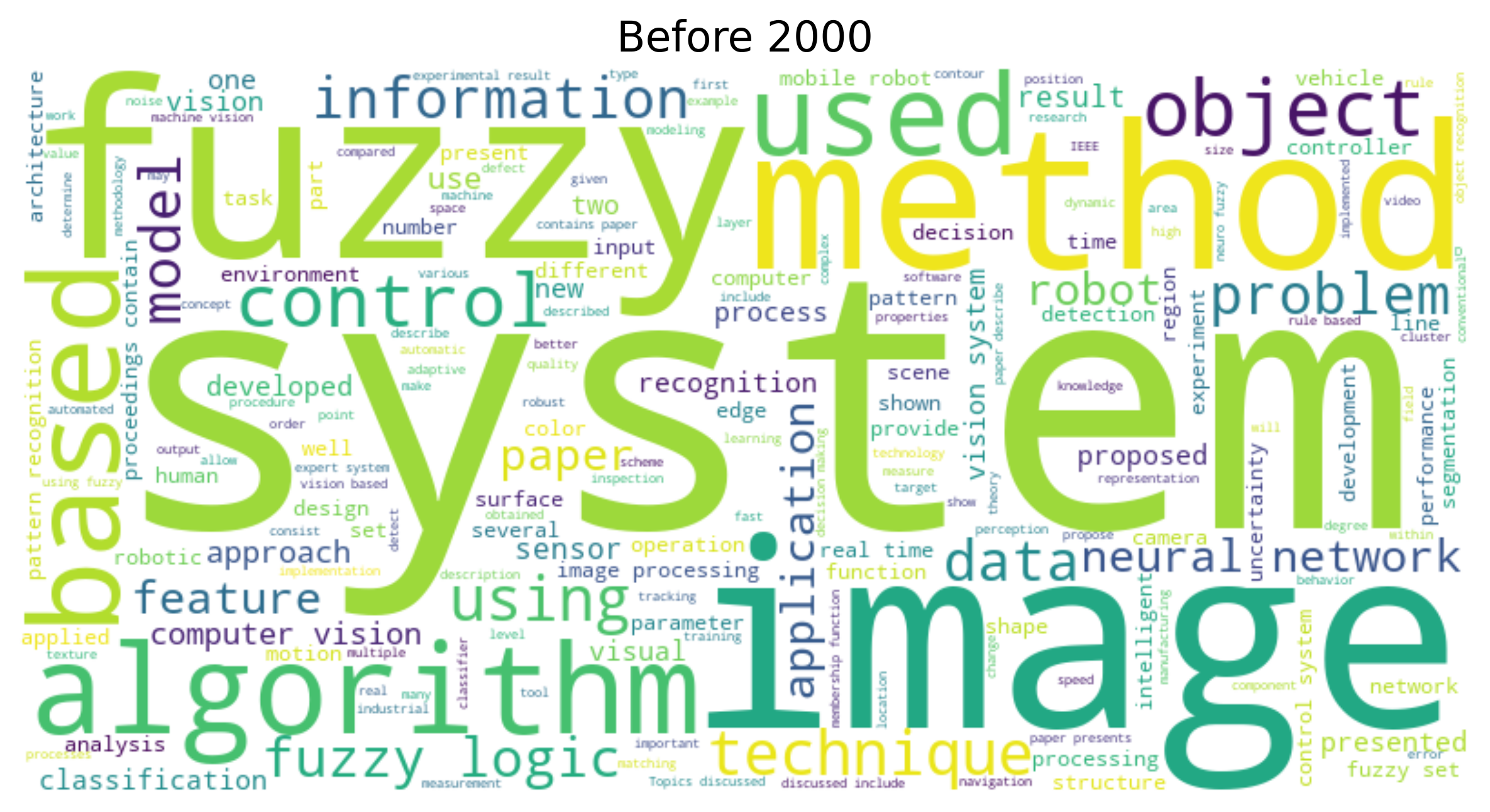}
        \caption{Before 2000}
    \end{subfigure}%
    \begin{subfigure}[t]{0.48\linewidth}
        \centering
        \includegraphics[width=\linewidth]{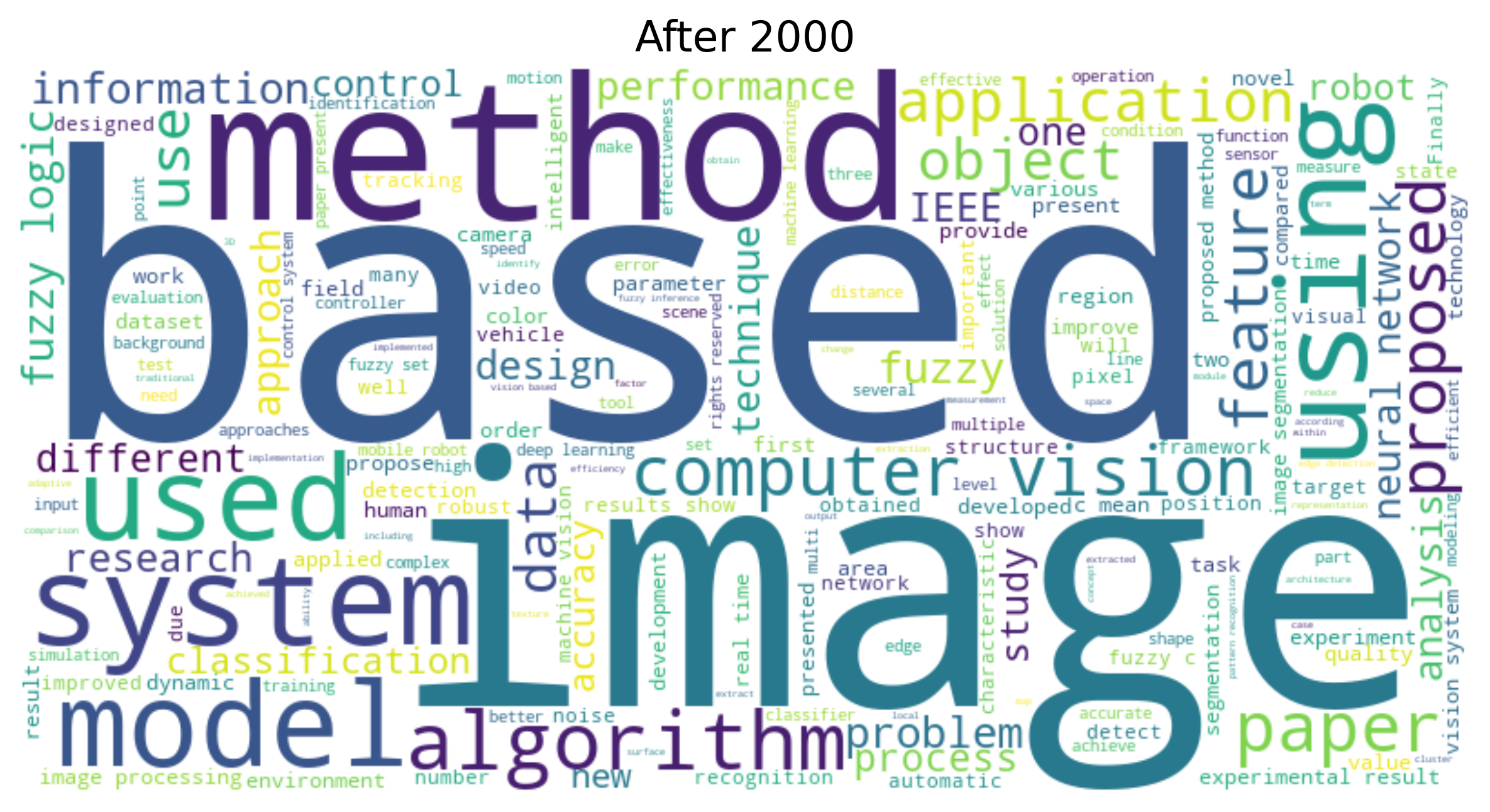}
        \caption{After 2000}
    \end{subfigure}
    \caption{Word Clouds of abstracts from publications on fuzzy computer vision in different time periods.}
    \label{abstract_wordcloud}
\end{figure*}

% \textcolor{purple}{we added it to Methodology, 1st paragraph must be replaced with 1 sentence}
We conducted a bibliometric analysis of publications on fuzzy computer vision. The dataset included 4,305 documents from Scopus published between 1973 and 2025. 

% The following are the key findings of our study. Table \ref{tab:top_publishers} presents information about the number of papers on fuzzy computer vision by publishers. IEEE emerges to be the leading publisher in the topic (961), followed by Springer (494) and Elsevier (248).  

Fig. \ref{fig_cloud} illustrates a word cloud of keywords from the analyzed papers. Besides the primary terms \textit{computer vision} and \textit{fuzzy logic}, notable keywords include \textit{neural network}, \textit{machine vision}, \textit{image processing}, \textit{recognition}, and \textit{deep learning}. Furthermore, terms such as \textit{fuzzy control}, \textit{feature extraction}, \textit{clustering}, and \textit{edge detection} indicate the frequent use of \textit{fuzzy logic} and \textit{machine learning}.
The network graph in Fig. \ref{fig_keyw_net} illustrates the co-occurrence of keywords within the collected dataset. Each node represents a distinct keyword, while the edges connecting them signify instances where the corresponding keywords appear together within the same article. In addition, the radius of the nodes indicates the frequency of usage of the respective keyword: \textit{fuzzy logic} is a larger node, suggesting it is among the most frequently used keywords in the dataset. Notably, \textit{computer vision}, \textit{machine vision}, \textit{deep learning}, \textit{image segmentation}, and \textit{image processing}, represented by average-sized nodes, are also frequently used, highlighting their significance as core concepts within the dataset, also illustrated in Figure \ref{fig_cloud} above. Moreover, the thickness of the edges denotes the strength of co-occurrence, with thicker edges representing more frequent co-occurrences between keywords. \textit{Fuzzy logic}, \textit{computer vision}, \textit{image processing}, and \textit{machine vision} emerge as the primary co-occurring keywords.
Fig. \ref{abstract_wordcloud} demonstrates word clouds of abstracts from publications on fuzzy computer vision. Comparison of word clouds from different time periods can show us how words like \textit{system}, \textit{fuzzy} and \textit{fuzzy logic} are used less in the current millennia and replaced with terms as \textit{model} and \textit{computer vision}.
Next, Fig. \ref{fig_trend_cite} illustrates the trends in publications and citations over time. The number of publications has steadily increased since the late 1980s. Citations show a sharp rise in the early 1990s, with two highly cited articles: \cite{Xie_85677} (1991) and \cite{Krishnapuram_227387} (1993) standing out. In general, the trend reflects the growing research output and its increasing citation over time. Fig. \ref{fig:overall_top10_key} presents the trends of keywords in publications from 1980 to 2025, illustrating their fluctuations over time.

\section{Conclusion}
\label{sec:conclusion}
The review demonstrated the significance of fuzzy logic in modern computer vision. The main advantage of using fuzzy logic in computer vision algorithms is its ability to handle uncertainty. The incorporation of fuzzy logic improves low-level feature processing, which forms the foundation of computer vision.  Unlike deep learning, which operates as a "black box," fuzzy systems provide transparent, rule-based reasoning, enhancing trust and interpretability. Additionally, fuzzy methods require fewer labeled samples, making them advantageous in scenarios with limited or noisy data. These properties make fuzzy logic a valuable tool for tasks such as segmentation, object recognition, and scene understanding, where uncertainty is inherent.

However, fuzzy logic presents several limitations, including difficulties in defining membership functions and constructing fuzzy rules, as it requires experts in relevant fields.
% to develop a qualitative system. This leads to another limitation: Approaching more accurate results with an uncertain system necessitates processing as much input data as possible, which exponentially increases the number of fuzzy rules and memberships. Due to these limitations, fuzzy logic is usually used in pairs with other machine learning algorithms, which is a common trend now.

% ... As for future challenges, it is...
% Stressing open research questions...

% \colorbox{Goldenrod}{10. Main message }

% \colorbox{Goldenrod}{11. Main findings }

% \colorbox{Goldenrod}{12. Study outcomes }

% \colorbox{Goldenrod}{13. Study limitations }

% \colorbox{Goldenrod}{14. Future works }

% Main message, ping.

% Summarize the development 

% Suggest future work

% Implications.
% \begin{table}[ht]
%  \centering
%  \makebox[\textwidth][c]{\includegraphics[width=1\textwidth]{Figures/Table_I.jpg}}%
%  \caption{Tables numbered in roman numerals}
%  \label{tab-II}
% \end{table}

% \textit{Acknowledgments} \\Mention all external funding sources in the acknowledgements.\\

% \textit{Disclosure statement:} \\No potential conflict of interest is reported by the authors
\section*{Acknowledgment}
This research has been funded by the Science Committee of the Ministry of Science and Higher Education of the Republic of Kazakhstan (Grant No. AP22786412)

\bibliographystyle{IEEEtran}
\bibliography{Bibliography}
\end{document}